\pdfoutput=1

\documentclass[11pt]{article}

\usepackage[preprint]{acl}

\usepackage{times}
\usepackage{latexsym}

\usepackage[T1]{fontenc}

\usepackage[utf8]{inputenc}

\usepackage{microtype}

\usepackage{inconsolata}

\usepackage{graphicx}

\usepackage{algorithmic,algorithm}
\renewcommand{\algorithmiccomment}[1]{\bgroup\hfill$\triangleright$~#1\egroup}
\usepackage[linguistics]{forest} 
\usepackage{synttree}
\usepackage{subcaption}
\usepackage{tikz-dependency}
\usepackage[cjk]{kotex}

\usepackage{amsmath}

\DeclareFontFamily{U}{mathb}{}
\DeclareFontShape{U}{mathb}{m}{n}{
  <-5.5> mathb5
  <5.5-6.5> mathb6
  <6.5-7.5> mathb7
  <7.5-8.5> mathb8
  <8.5-9.5> mathb9
  <9.5-11.5> mathb10
  <11.5-> mathb12
}{}
\DeclareSymbolFont{mathb}{U}{mathb}{m}{n}
\DeclareMathSymbol{\drsh}{3}{mathb}{"EB}
\usepackage{amssymb}

\usepackage{amsthm}

\newtheorem{property}{Property}
\newenvironment{proofsketch}{\par\noindent\textit{Proof sketch}\ }{\hfill$\square$\par}

\newenvironment{interpretation}{\par\noindent\textit{Interpretation}\ }{\hfill$\square$\par}

\usepackage{booktabs}

\usepackage{langsci-gb4e}

\title{
A Formal Framework for Fluency-based Multi-Reference Evaluation in Grammatical Error Correction
}

\author{
Eitan Klinger$^1$\thanks{Equally contributed authors.\\
$^{\dagger}$Corresponding authors: Mengyang Qiu and Jungyeul Park.}~~  Zihao Huang$^{2*}$~~  Tran Minh Nguyen$^{2*}$~~ Emma Jayeon Park$^3$\\
{\bf Yige Chen}$^4$~~ {\bf Yang Gu}$^2$~~ {\bf Qingyu Gao}$^2$~~ {\bf Siliang Liu}$^2$~~ {\bf Mengyang Qiu}$^{2,5\dagger}$~~ {\bf Jungyeul Park}$^{2,6\dagger}$\\
$^1$The University of British Columbia, Canada~~ 
$^2$Open Writing Evaluation, France\\
$^{3}$Université de Rennes, France~~
$^{4}$The Chinese University of Hong Kong, Hong Kong\\
$^{5}$Trent University, Canada~~
$^{5}$KAIST, South Korea\\
{\tt \url{http://open-writing-evaluation.github.io}}
}

\begin{document}
\maketitle

\begin{abstract}
Evaluating grammatical error correction requires metrics that reflect the diversity of valid human corrections rather than privileging a single reference.  
Existing frameworks, largely edit-based and English-centric, rely on rigid alignments between system and reference edits, limiting their applicability in multilingual and generative settings.  
This paper introduces a formal framework for \textit{fluency-based multi-reference evaluation}, framing $n$-gram similarity as an aggregation problem over multiple legitimate corrections.  
Within this formulation, we instantiate GLEU through four aggregation strategies—\textsc{select-best}, \textsc{simple-average}, \textsc{weighted-average}, and \textsc{merged-counts}—and analyze their properties of boundedness, monotonicity, and sensitivity to reference variation.  
Empirical results on Czech, Estonian, Ukrainian, and Chinese corpora show that these strategies capture complementary aspects of fluency and coverage.  
The framework unifies multi-reference evaluation into a principled, fluency-oriented approach that incorporates linguistic diversity without penalizing legitimate variation.
\end{abstract}

\section{Introduction}
Evaluating grammatical error correction (GEC) remains challenging because a learner sentence rarely admits a single unique correction.
Multiple valid rewrites may differ in lexical choice, phrasing, or stylistic preference, yet all improve the fluency and grammaticality of the original text.
This multiplicity raises a fundamental question: how should evaluation be defined when system outputs align with some, but not all, legitimate human corrections?

Current GEC evaluation practices are dominated by single-reference and edit-based frameworks such as \texttt{M$^2$} \citep{dahlmeier-ng-2012-better} and \texttt{ERRANT} \citep{bryant-etal-2017-automatic}.
These metrics rely on span-level edit matching between system and reference outputs.
While effective for English, they presuppose a one-to-one correspondence between errors and edits, which becomes unreliable in multilingual or generative contexts where equally valid corrections diverge in surface realization.
Dependence on a single canonical reference thus underestimates the quality of legitimate outputs that differ from the chosen gold correction.

Fluency-oriented metrics such as GLEU \citep{napoles-etal-2015-ground,napoles-etal-2016-gleu} partially address this issue by evaluating $n$-gram overlap between system and reference sentences, thereby capturing fluency rather than edit alignment.
However, standard implementations assume either a single reference or a naïve averaging across multiple references.
Such schemes lack theoretical grounding: they do not specify what formal properties multi-reference evaluation should satisfy, nor how aggregation choices influence fairness, stability, or linguistic coverage.
In practice, the absence of a principled formulation obscures the interpretation of results across tasks, languages, and reference sets.

Multi-reference evaluation is essential for representing the inherent variability of human language.
A single reference captures only one realization among many possible fluent variants.
Aggregating over multiple references mitigates spurious penalization, improves robustness to stylistic and structural diversity, and enhances cross-linguistic comparability across languages that differ in morphology, word order, or segmentation.
A principled multi-reference formulation therefore moves evaluation closer to the intuitions underlying human judgment of acceptability.

This paper develops a theoretical framework for fluency-based multi-reference evaluation.
We define evaluation as an aggregation function over sets of valid linguistic variants, generalizing beyond single-reference comparison.
Within this framework, we instantiate four aggregation strategies--\textsc{select-best}, \textsc{simple-average}, \textsc{weighted-average}, and \textsc{merged-counts}--and analyze their mathematical properties and empirical behavior.
Fluency-based metrics such as GLEU naturally emerge as specific realizations within this broader model rather than as isolated heuristics.

Our contributions are as follows.
(1) We propose a general mathematical framework for multi-reference evaluation, formalizing it as an aggregation function over sets of valid linguistic variants.
(2) We derive theoretical properties such as boundedness, monotonicity, and non-penalization that any valid aggregation function should satisfy.
(3) We show that fluency-based metrics such as GLEU instantiate a subclass of this formulation, enabling principled comparison among aggregation strategies.
(4) We empirically verify these properties across typologically diverse datasets, demonstrating the framework’s robustness and cross-linguistic validity.

By unifying existing approaches to GEC evaluation, this work establishes a formal foundation for reasoning about how multiple human references can be integrated into a single, interpretable score.
The proposed framework thus advances GEC evaluation from heuristic multi-reference averaging toward a principled, fluency-oriented theory of linguistic adequacy.

\section{Fluency-based multi-reference evaluation}

This section presents a fluency-oriented framework for evaluating GEC.  
Unlike edit-based approaches that depend on token-level alignments, fluency-based evaluation measures surface similarity through $n$-gram overlap, offering a more flexible and language-agnostic assessment of correction quality.  
We formalize its core principles and motivate the extension to multi-reference settings.

\subsection{From edit-based to fluency-based evaluation}

Edit-based evaluation metrics, such as \texttt{M$^2$} \citep{dahlmeier-ng-2012-better} and \texttt{ERRANT} \citep{bryant-etal-2017-automatic}, compare predicted and reference edits extracted through explicit alignments between the source sentence and its corrected form. This approach has clear interpretability advantages, but it also suffers from structural brittleness \citep{wang-etal-2025-refined}. Small differences in segmentation, word order, or span definition can cause large discrepancies in scores. Moreover, edit-based evaluation assumes a consistent tokenization scheme and predefined error taxonomy, making cross-linguistic comparison difficult. In multilingual or generative contexts, where alternative corrections may differ lexically yet remain equally fluent, span-level alignment becomes unreliable.

Fluency-based evaluation provides a complementary view of correction quality by assessing the surface well-formedness of the output rather than its exact edit structure. The most widely used fluency-oriented metric is \texttt{GLEU} \citep{napoles-etal-2015-ground,napoles-etal-2016-gleu}, a monolingual adaptation of BLEU \citep{papineni-etal-2002-bleu}. Unlike BLEU, which measures overlap between a system translation and its references, GLEU incorporates the source sentence to penalize uncorrected errors and reward successful revisions. Specifically, $n$-grams that appear in the reference but not in the source are rewarded, while those that persist from the source but should have been corrected are penalized. This design aligns with the goal of GEC systems: to make minimal, targeted edits that improve grammaticality and fluency.

Because GLEU operates at the level of $n$-gram overlap, it is less sensitive to the precise form of edits and more tolerant of lexical or syntactic variation. Two corrected sentences that differ in word choice or phrasing but share fluent surface patterns can achieve similar scores, reflecting human judgments of acceptability. This property makes GLEU particularly well-suited to multilingual evaluation, where differences in morphology and segmentation can obscure token-level correspondences.

Despite these advantages, standard implementations of GLEU already support multiple references, but only through a simple averaging scheme that treats each reference independently. This approach does not fully capture the diversity of human corrections, as it ignores interactions among references and their varying linguistic plausibility. In practice, multiple valid rewrites often exist for a single learner sentence, differing in both structure and style. A more principled multi-reference formulation is therefore needed to represent this variability and provide a robust, fluency-oriented measure of correction quality. In the following subsection, we formalize this extension and introduce four aggregation strategies that define how multiple references can be integrated into a single evaluation score.

\subsection{GLEU under multiple references}

The GLEU metric \citep{napoles-etal-2015-ground,napoles-etal-2016-gleu} was proposed as a fluency-oriented alternative to edit-based metrics for grammatical error correction. It adapts the logic of BLEU \citep{papineni-etal-2002-bleu} to the monolingual rewriting setting by comparing the system hypothesis $H$ not only to the reference correction $R$ but also to the original uncorrected source $S$. In contrast to BLEU, which rewards all $n$-gram matches between $H$ and $R$, GLEU rewards only those $n$-grams that should have been changed (i.e., appear in $R$ but not in $S$) and penalizes those that remain erroneous (i.e., appear in $S$ but not in $R$). This design aligns the metric with the goal of grammatical error correction: to make minimal but meaningful edits that improve fluency and correctness.

Formally, for a hypothesis $H$, a source $S$, and a reference correction $R$, GLEU computes modified $n$-gram counts that jointly reward correct edits and penalize uncorrected errors.  
For each $n$-gram $g$, the effective matched count is defined as
\[
\begin{aligned}
 & \mathrm{count}'(g)\\
=\quad &\max \Big(
  \min(\mathrm{C}_H(g), \mathrm{C}_R(g)) \\
& - \min(\mathrm{C}_H(g), \max(0, \mathrm{C}_S(g) - \mathrm{C}_R(g))),
  0
\Big)
\end{aligned}
\]
where $\mathrm{C}_X(g)$ denotes the frequency of $g$ in sequence $X$.  
This formulation ensures that $n$-grams shared with the reference contribute positively, while those shared with the source but absent from the reference contribute negatively.

The modified $n$-gram precision for order $n$ is then
\[
P_n =
\frac{
\sum_{g \in \text{ngrams}_n(H)} \mathrm{count}'(g)
}{
\sum_{g \in \text{ngrams}_n(H)} \mathrm{C}_H(g)
}
\]
and the final GLEU score is computed using the standard BLEU brevity penalty (BP) and geometric mean:
\[
\begin{aligned}
& \text{GLEU}(S,H,R)\\
= \quad & \text{BP} \cdot
\mathrm{exp}\left(
\sum_{n=1}^{N} w_n \log P_n
\right) \\
&\quad \text{where } w_n = \tfrac{1}{N},\; N = 4
\end{aligned}
\]
The brevity penalty is identical to that of BLEU:
\[
\text{BP} =
\begin{cases}
1, & c > r\\
e^{(1 - r/c)}, & c \le r
\end{cases}
\]
where $c$ and $r$ denote the lengths of the hypothesis and reference, respectively.

This formulation preserves BLEU’s fluency sensitivity while incorporating explicit penalties for uncorrected source fragments, yielding a metric that directly reflects the objectives of grammatical error correction.

Extending GLEU to multi-reference evaluation is non-trivial. Each reference correction $R_i$ represents a distinct valid realization of the same intended meaning, differing in lexical choice, word order, or phrasing. As a result, the corresponding sets of valid $n$-grams may overlap only partially across references. Simply merging all reference $n$-grams can inflate frequency counts and blur stylistic variation, while naive averaging may underrepresent infrequent yet valid corrections. A principled aggregation mechanism is therefore required to integrate multiple references without distorting their linguistic diversity.

We define the general multi-reference formulation of GLEU as:
\[
\begin{aligned}
& \text{GLEU}(S,H,\{R_i\}) \\
= & \quad  \text{Aggregation}\big( \\
& \quad\quad \text{GLEU}(S,H,R_1), \dots, \text{GLEU}(S,H,R_n)
\big)
\end{aligned}
\]
where \textit{Aggregation} determines how individual reference scores are combined into a single fluency-based evaluation. Different aggregation strategies reflect distinct evaluation philosophies: \textsc{select-best} privileges the reference most similar to the hypothesis, \textsc{simple-average} assigns equal weight to all references, \textsc{weighted-average} emphasizes prototypical or majority-like corrections, and \textsc{merged-$n$-grams} maximizes recall by rewarding any overlap with human-validated expressions.

These aggregation methods operationalize how multi-reference fluency evaluation balances precision, diversity, and human acceptability, forming the empirical and theoretical foundation for the analyses presented in the following section.

\section{Aggregation strategies for multi-reference GLEU}

This section presents the aggregation strategies used to extend fluency-based evaluation to multi-reference grammatical error correction. 
When multiple valid references exist, the challenge lies in determining how to integrate these distinct yet legitimate corrections into a single, interpretable score. We describe the general evaluation pipeline and formalize four aggregation strategies that define how individual reference scores are combined.

\subsection{Evaluation pipeline}

The multi-reference GLEU evaluation pipeline proceeds in three steps:
(1) Compute the GLEU score between the hypothesis $H$ and each reference $R_i$, given the original source $S$.  
(2) Combine the resulting set of scores $\{\text{GLEU}(S,H,R_i)\}_{i=1}^{n}$ using one of the four aggregation strategies described below.  
(3) Report the corpus-level GLEU score by averaging sentence-level results across all test samples.

This procedure preserves the interpretability of GLEU while enabling systematic comparison across multiple valid human corrections. The choice of aggregation strategy determines how linguistic diversity and stylistic variation among references influence the final evaluation outcome.

\subsection{Multi-reference aggregation strategies}

This subsection defines four aggregation strategies for computing GLEU under multiple references.  
Each strategy reflects a distinct evaluation philosophy regarding how to balance precision, diversity, and stylistic variation among valid human corrections.  
Formally, all strategies instantiate a mapping
\[
\begin{aligned}
\Phi : 
& \; (\text{GLEU}(S,H,R_1),\dots,\text{GLEU}(S,H,R_n)) \\
& \mapsto [0,1]
\end{aligned}
\]
subject to properties such as boundedness, monotonicity, and non-penalization with respect to the reference set.

\subsubsection{Select-best}

The \textsc{select-best} strategy evaluates the hypothesis against each reference independently and retains the highest GLEU score, providing an oracle upper bound on achievable fluency.
\[
\begin{aligned}
& \text{GLEU}_{\textsc{select-best}}(S,H,\{R_i\}) \\
= & \quad \max_{i} \, \text{GLEU}(S,H,R_i)
\end{aligned}
\]
It is precision-oriented, rewarding systems that produce outputs closely matching at least one reference, while disregarding variation across others.  
Conceptually, it favors stylistic proximity and minimizes penalties for divergence from less similar but valid alternatives.  
Theoretically, it satisfies \textit{boundedness} (scores remain within [0,1]) and \textit{monotonicity} (adding a new reference cannot decrease the score), but violates \textit{fairness}, since all but the best-scoring reference are ignored.  
This corresponds to an \emph{existential} interpretation of correctness: the hypothesis is valid if there exists at least one matching reference.

\subsubsection{Simple average}

The \textsc{simple-average} strategy assigns equal weight to all references, representing a uniform aggregation function that captures the diversity of acceptable corrections without privileging any particular style or phrasing.  
\[
\begin{aligned}
& \text{GLEU}_{\textsc{simple-avg}}(S,H,\{R_i\}) \\
= & \quad \frac{1}{n} \sum_{i=1}^{n} \text{GLEU}(S,H,R_i)
\end{aligned}
\]
It is liberal in interpretation, emphasizing representational balance over selectivity and reflecting an inclusive, fairness-oriented perspective.  
Theoretically, it satisfies \textit{boundedness}, \textit{symmetry}, and \textit{fairness}, since every reference contributes equally.  
However, it may violate \textit{monotonicity}: adding low-scoring references can decrease the overall average, potentially penalizing diversity.  
This strategy corresponds to a \emph{collective} interpretation of correctness, emphasizing inclusivity while assuming equal plausibility among human variants.

\subsubsection{Weighted average}

The \textsc{weighted-average} strategy defines a softly monotonic aggregation function: as $\tau$ increases, it converges to \textsc{select-best}, and as $\tau \to 0$, to \textsc{simple-average}.  
\[
w_i =
\frac{\mathrm{exp}(\tau \cdot \text{GLEU}(S,H,R_i))}
{\sum_{j=1}^{n} \mathrm{exp}(\tau \cdot \text{GLEU}(S,H,R_j))},
\quad \tau > 0
\]
Here, $\tau$ controls the sharpness of weighting, with larger values favoring the most similar references.  
The final multi-reference score is defined as
\[
\begin{aligned}
& \text{GLEU}_{\textsc{weighted-avg}}(S,H,\{R_i\}) \\
= & \quad \sum_{i=1}^{n} w_i \, \text{GLEU}(S,H,R_i)
\end{aligned}
\]
This strategy models a graded notion of acceptability: frequent or prototypical corrections exert stronger influence, while rare variants are still recognized.  
It balances inclusivity and selectivity, aligning with human tendencies to judge correctness probabilistically rather than categorically.  
Theoretically, it satisfies \textit{boundedness} and \textit{non-penalization}, providing a continuous trade-off between selectivity and inclusivity within a probabilistic interpretation of fluency and human preference aggregation.

\subsubsection{Merged $n$-grams}

The \textsc{merged-$n$-grams} strategy performs aggregation at the representation level rather than at the score level, combining all $n$-grams from the reference set $\{R_i\}$ into a single multiset:
\[
R_{\cup} = \bigcup_{i=1}^{n} \text{ngrams}(R_i)
\]
GLEU is then computed once using $R_{\cup}$ as the composite reference:
\[
\text{GLEU}_{\textsc{merged}}(S,H,\{R_i\})
= \text{GLEU}(S,H,R_{\cup})
\]
This approach prioritizes coverage and recall by rewarding overlap with any human-validated expression.  
It tolerates stylistic variation and paraphrasing, reflecting an evaluation philosophy centered on inclusivity and linguistic flexibility rather than strict matching.  
Formally, it guarantees \textit{monotonicity} (since $R_{\cup}$ expands with each added reference) and \textit{non-penalization}, but not \textit{fairness}, as longer or denser references contribute proportionally more $n$-grams.  
This corresponds to a \emph{union-based} interpretation of correctness that prioritizes recall and linguistic coverage over stylistic precision.

\paragraph{Comparative interpretation}
Each strategy corresponds to a distinct aggregation operator in the formal space of evaluation functions:
\textsc{select-best} realizes the existential extreme,
\textsc{merged-$n$-grams} the universal one,
while \textsc{simple-average} and \textsc{weighted-average} provide interpretable trade-offs between inclusivity and selectivity.
These perspectives clarify how different aggregation principles balance precision, fairness, and coverage in the fluency-based multi-reference evaluation framework.

\paragraph{Mathematical formulation.}
A general multi-reference fluency metric can be defined as an aggregation operator over sentence-level GLEU scores that satisfies key properties such as boundedness, monotonicity, and symmetry.  
Each of the four strategies described above corresponds to a distinct functional form of this operator, defining a continuum between precision-oriented and recall-oriented evaluation.  
A full formalization, including proofs of these properties, is provided in Appendix~\ref{appendix:math-formulation}.

\subsection{Analytical properties}

The four aggregation strategies introduced above exhibit distinct theoretical behaviors that determine how multi-reference GLEU responds to variation in reference diversity and stylistic overlap.  
Table~\ref{tab:gleu-properties} summarizes their core analytical properties, highlighting how each strategy positions itself along the precision–recall and inclusivity–selectivity spectra.

\begin{table}[!ht]
\centering
\resizebox{.48\textwidth}{!}{
\footnotesize
\begin{tabular}{lcccc}
\toprule
 & \textsc{select} & \textsc{simple} & \textsc{weighted} & \textsc{merged} \\
 & \textsc{best} & \textsc{avg} & \textsc{avg} & \textsc{$n$-grams} \\
\midrule
Recall dominance             &--&--&--& \checkmark \\
Oracle upper bound           & \checkmark &--&--&--\\
Permutation invariance       &--& \checkmark &--& \checkmark \\
Stability under segmentation &--&--&--& \checkmark \\
\bottomrule
\end{tabular}}
\caption{Analytical properties of multi-reference GLEU aggregation strategies.}
\label{tab:gleu-properties}
\end{table}

The \textsc{merged} strategy exhibits \textit{recall dominance}, since its unified $n$-gram pool captures every valid expression attested in the reference set, making it the most inclusive and recall-oriented formulation.  
Conversely, the \textsc{select-best} strategy represents an \textit{oracle upper bound}: by choosing the most favorable reference for each sentence, it yields the theoretical maximum achievable score.  
Both \textsc{simple-average} and \textsc{merged} GLEU are invariant to the ordering of references, whereas \textsc{weighted-average} is order-sensitive only through its weighting mechanism.  
Finally, the \textsc{merged} strategy remains stable under minor segmentation differences, as it operates over aggregated $n$-gram counts rather than one-to-one reference alignments.

These analytical distinctions mirror human judgment patterns.  
The \textsc{select-best} strategy approximates a strict grading style, rewarding only the closest match to a reference.  
The \textsc{merged} strategy reflects inclusive human evaluation, acknowledging all plausible corrections as acceptable.  
Between these extremes, the \textsc{weighted-average} approach captures graded human acceptability, rewarding outputs that align with prototypical or majority forms while still assigning partial credit to less frequent but valid alternatives.

\section{Experiments and results} \label{results}

This section presents the empirical results of fluency-based multi-reference evaluation using GLEU and its four aggregation strategies.  
We first describe the datasets and experimental setup, emphasizing development sets that contain multiple human corrections.  
We then outline our essay-level prompting and evaluation methodology using a large language model, followed by results obtained under single- and multi-reference fluency evaluation.  
Finally, we analyze the effects of reference multiplicity, aggregation strategy, and cross-lingual variation on fluency-based system assessment.

\subsection{Datasets}

Table~\ref{dev-stats} summarizes the multilingual development sets \citep{masciolini-etal-2025-multigec} used in our experiments.  
Each corpus includes up to two human reference corrections per essay, enabling us to analyze how fluency-based evaluation behaves across varying levels of reference diversity.
Czech provides overlapping L2 corrections, Estonian challenges evaluation with rich morphology, and Ukrainian, with full multi-reference annotation, serves as the main benchmark.\footnote{\url{https://github.com/spraakbanken/multigec-2025}}
These datasets collectively cover a wide range of linguistic and annotation conditions for multi-reference evaluation.

\begin{table}[!ht]
\centering
{
\footnotesize
\begin{tabular}{r ccc}\toprule
 & \textsc{orig} & \textsc{ref 1} & \textsc{ref 2} \\
\midrule
Estonian--EIC           & 26   & 26   & 26 \\
Estonian--EKIL2         & 150  & 150  & 150 \\
Czech--SecLearn         & 173  & 173  & 97 \\
Czech--NatForm          & 88   & 88   & 47 \\
Czech--NatWebInf        & 1291 & 1291 & 687 \\
Czech--Romani           & 179  & 179  & 84 \\
Ukrainian--UA-GEC       & 87   & 87   & 87 \\
\bottomrule
\end{tabular}
}
\caption{Development set sizes by language. Each corpus provides at least two human reference corrections per essay (fully or partially).}
\label{dev-stats}
\end{table}

The use of multi-reference corpora is essential for evaluating fluency-based metrics such as GLEU.  
Since different human corrections often vary in lexical and syntactic realization, single-reference scoring can underestimate valid model outputs that align with non-selected references.  
Multi-reference evaluation therefore provides a more realistic and robust reflection of linguistic acceptability at the document level.  
Although this study does not directly correlate metric scores with human judgments, the inclusion of typologically diverse datasets offers an empirical basis for assessing the framework’s cross-linguistic robustness.  
We leave large-scale validation against human ratings to future work.

\subsection{Prompting and evaluation methodology}

We evaluate grammatical error correction performance using a large language model without task-specific fine-tuning.\footnote{\url{https://huggingface.co/deepseek-ai/DeepSeek-V3}}  
To approximate real-world writing-assistance scenarios, we adopt an \emph{essay-level} correction setup rather than isolated sentence-level prompts.  
This design captures broader fluency phenomena such as discourse coherence, lexical balance, and stylistic consistency across paragraphs.

Each essay is introduced with the instruction:  
\textit{``Correct the following essay to standard \{Czech|Estonian|Ukrainian\} language.''}  
The model then outputs one fully corrected essay.  
Deterministic decoding (temperature $=0.0$, top-$p=1.0$) ensures reproducibility.  
Outputs are normalized for whitespace and punctuation before computing $n$-gram statistics.

Evaluation is performed at the \emph{essay level} without sentence alignment.  
For each essay, GLEU is computed between the model hypothesis $H$, the uncorrected source $S$, and all available human reference essays $\{R_1, R_2\}$.  
Several $R_2$ subsets are only partially annotated and cover a limited portion of each corpus.  
The resulting GLEU scores are combined using one of the four aggregation strategies--\textsc{select-best}, \textsc{simple-average}, \textsc{weighted-average}, or \textsc{merged-$n$-grams}.  
The final corpus-level result is the mean of these essay-level scores.

As the primary objective of this study is to examine the behavior of fluency-based metrics rather than to benchmark GEC systems, we use a single hypothesis per essay.  
Prompting thus serves as a controlled generation method that ensures consistency across languages and provides a reproducible setting for analyzing the properties of multi-reference evaluation.

\subsection{Results}

Table~\ref{gec-results-fluency} presents the results of fluency-based multi-reference evaluation across all languages and aggregation strategies, compared with the single-reference baseline. Scores are computed at the essay level and aggregated to the corpus level, ensuring consistency between generation and evaluation granularity.

\begin{table*}[!th]
\centering
\small
\begin{tabular}{r cccccc}
\toprule
{} & {Select-best} & {Avg} & {Weighted} & {Merged} & {Single-ref} & $\boldsymbol{\Delta}$ (\textit{Merged--Single}) \\ 
\midrule
Estonian--EIC & 0.5304 & 0.4997 & 0.5139 & {0.5609} & 0.5264 & +0.0345 \\
Estonian--EKIL2 & 0.4574 & 0.4520 & 0.4533 & {0.4584} & 0.4494 & +0.0090 \\
Czech--SecLearn & 0.5625 & 0.5383 & 0.5512 & {0.6034} & 0.5088 & +0.0946 \\
Czech--NatForm & 0.6808 & 0.6501 & 0.6701 & {0.7003} & 0.6187 & +0.0816 \\
Czech--NatWebInf & 0.5816 & 0.5563 & 0.5771 & {0.5861} & 0.5563 & +0.0298 \\
Czech--Romani & 0.5440 & 0.5168 & 0.5364 & {0.5649} & 0.5005 & +0.0644 \\
Ukrainian--UA-GEC & 0.6488 & 0.6119 & 0.6320 & {0.6932} & 0.6325 & +0.0607 \\
\bottomrule
\end{tabular}
\caption{Corpus-level GLEU scores for multi-reference evaluation compared to the single-reference baseline. The final column shows the improvement ($\Delta$) of merged multi-reference GLEU over the single-reference setting.}
\label{gec-results-fluency}
\end{table*}

Across all datasets, multi-reference evaluation consistently improves corpus-level GLEU relative to the single-reference baseline.  
The \textsc{merged-$n$-grams} strategy yields the highest scores in every corpus, with an average improvement of +0.06, confirming its recall-oriented behavior predicted by the analytical framework.  
Differences between \textsc{merged} and \textsc{simple-average} are statistically significant under paired bootstrap resampling ($p < 0.05$) for the Czech and Ukrainian datasets, indicating that the observed gains are not due to random variation.  
This consistency across typologically distinct languages supports the robustness of fluency-based multi-reference evaluation and motivates the detailed cross-linguistic analysis in Section~\ref{sec:crossling}.

\section{Cross-linguistic analysis of multi-reference evaluation} \label{sec:crossling}

This section provides an empirical interpretation of how multi-reference evaluation behaves across languages and datasets.  
Rather than comparing models or scoring variants, the analysis focuses on how reference aggregation interacts with corpus characteristics such as annotation density, correctional diversity, and segmentation granularity.  
Across both European and Chinese corpora, the effects of aggregation vary with reference distribution but tend to converge once reference coverage becomes sufficient.  
This convergence supports the cross-linguistic robustness of the proposed framework, showing that increased reference diversity systematically reduces penalization and enhances interpretability.

\subsection{Variation in multi-reference gains across corpora}

The impact of multi-reference aggregation differs markedly across the evaluated datasets.  
Czech corpora, especially SecLearn and NatForm, show the largest gains between the merged and single-reference settings ($\Delta$ of +0.09 and +0.08, respectively).  
These improvements suggest that Czech references exhibit greater heterogeneity in lexical and syntactic realization, allowing the merged $n$-gram pool to capture a broader range of acceptable expressions.  
This behavior aligns with the framework’s prediction that aggregation benefits increase with reference diversity.  

By contrast, Estonian and Ukrainian datasets display smaller improvements, reflecting either limited reference variation or lower annotation density per sentence.  
In Estonian, morphological complexity constrains the space of fluently interchangeable forms, while in Ukrainian the references are more consistently aligned, reducing the effect of aggregation.  
Across all corpora, multi-reference aggregation amplifies corpus-level recall while maintaining score stability in datasets with high annotation consistency, confirming that the proposed framework remains reliable under both high- and low-variance annotation conditions.

\subsection{Convergence and interpretability of aggregation strategies}

All corpora exhibit the same ordering of aggregation strategies—\textsc{merged} $>$ \textsc{select-best} $>$ \textsc{weighted} $>$ \textsc{average}—yet the magnitude of difference narrows as reference coverage increases.  
This convergence indicates that once reference diversity reaches a sufficient level, the influence of the aggregation function on overall GLEU becomes minimal.  
Beyond this point, additional references primarily serve to reduce penalization rather than to inflate scores.

This behavior can be explained by the increasing overlap among reference $n$-grams: as the set of references becomes more saturated, each aggregation function approximates the same empirical upper bound.  
\textsc{Select-best} and \textsc{weighted-average} converge toward \textsc{merged} because most plausible $n$-grams are already represented in at least one reference, reducing the effect of selection bias.  
Conversely, when references are sparse, aggregation choice has a stronger effect on recall and fairness.  

Thus, the interpretability of multi-reference evaluation lies not in absolute score comparison but in the stability of scores under expanding reference coverage.  
This property supports the framework’s intended goal: to evaluate systems in a way that reflects linguistic variability while maintaining consistent, cross-linguistic interpretability.

\subsection{Evaluating reference aggregation behavior in MuCGEC}

We analyze the development portion of the MuCGEC corpus \citep{zhang-etal-2022-mucgec}, which contains 1,137 sentences with up to seven independently annotated reference corrections.\footnote{\url{https://github.com/HillZhang1999/MuCGEC}}  
The reference distribution is highly uneven: 287 sentences have a single reference, 462 have two, 313 have three, and only a few contain four or more (62 with four, 10 with five, 2 with six, and 1 with seven).  
On average, each sentence has 2.16 references.  
Since MuCGEC does not include explicit word boundaries, all GLEU computations are performed at the character $n$-gram level.

\begin{figure}[!ht]
\centering
\includegraphics[width=0.98\linewidth]{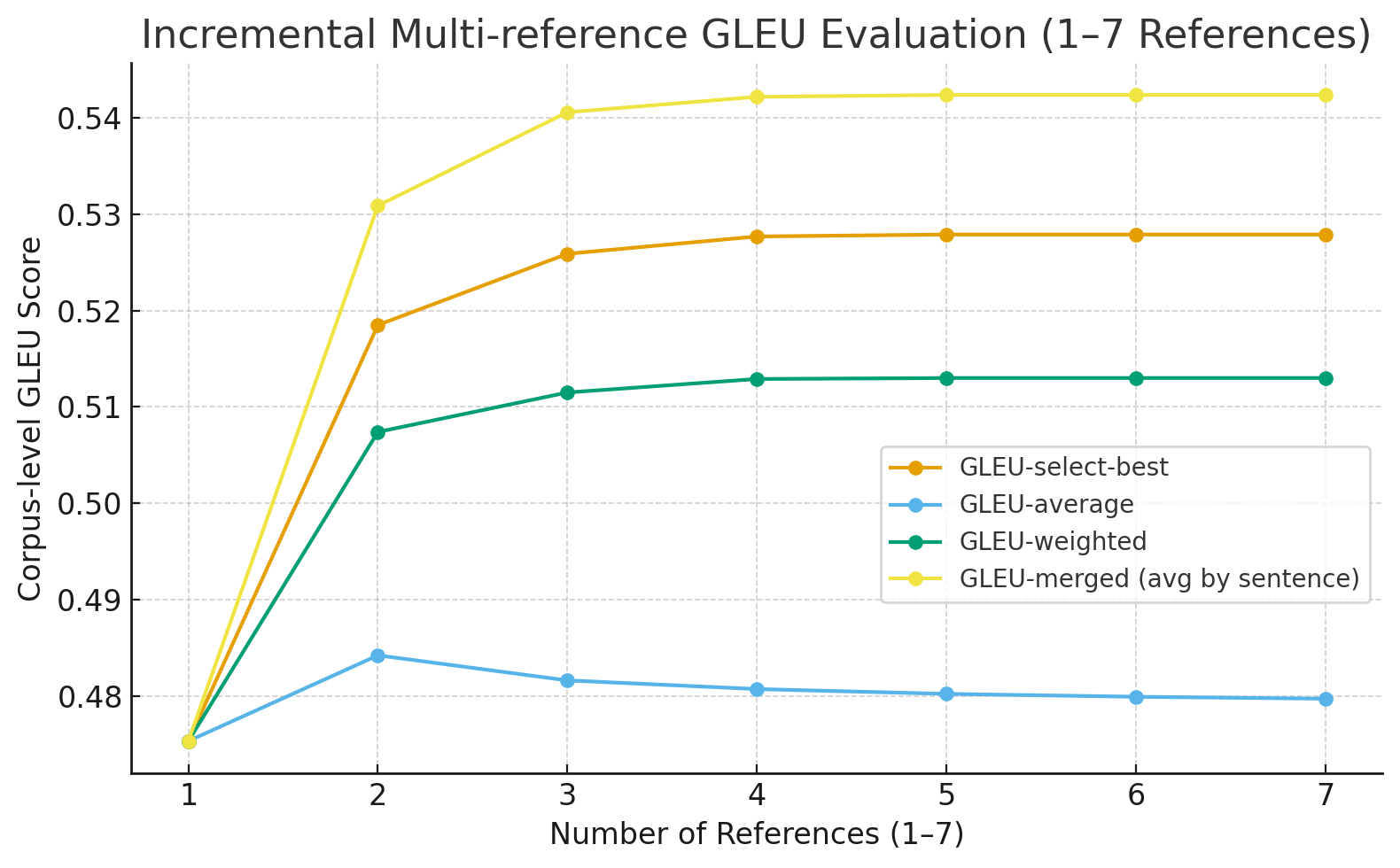}
\caption{Incremental multi-reference evaluation on the MuCGEC dataset using corpus-level GLEU.  
The \textsc{merged} variant steadily increases as additional references are included, while \textsc{select-best} plateaus and \textsc{average} remains lower.  
This illustrates how reference aggregation enhances recall without sacrificing stability.}
\label{mucgec-gleu}
\end{figure}

Figure~\ref{mucgec-gleu} shows corpus-level GLEU scores as the number of available references increases from one to seven.  
Initially, all variants (\textsc{select-best}, \textsc{average}, \textsc{weighted}, and \textsc{merged}) coincide at 0.4753, but as more references are added, they diverge and stabilize after about three to four references.  
This stabilization reflects the sparsity of sentences with higher reference counts, as new annotations beyond that point contribute little to the corpus-level average.

The goal of this analysis is not to identify the highest-scoring variant but to examine how aggregation affects evaluation dynamics as reference coverage expands.  
The upward trajectory of \textsc{merged} empirically confirms its recall-dominant and monotonic behavior, while the plateau of \textsc{select-best} reflects its bounded precision-oriented design.  
The \textsc{weighted} and \textsc{average} strategies converge between these extremes, validating their role as graded aggregation functions that trade off inclusivity and selectivity.  
Together, these findings demonstrate that reference diversity systematically improves fairness and interpretability, extending the robustness of fluency-based multi-reference evaluation to typologically distant languages such as Chinese.

\section{Conclusion and future perspectives}

This work presented a formal framework for evaluating grammatical error correction systems under multi-reference conditions, where each human correction is treated as a valid realization rather than a competing alternative.
By framing $n$-gram–based fluency evaluation as an aggregation problem, we unified existing approaches and clarified the mathematical properties—boundedness, monotonicity, and non-penalization—that define a sound multi-reference metric.
Empirical analyses across typologically diverse corpora further demonstrated that different aggregation strategies capture complementary dimensions of fluency and coverage, highlighting the importance of linguistic diversity in evaluation.

The proposed formulation thus moves GEC evaluation from heuristic averaging toward a principled, fluency-oriented perspective that accommodates multiple legitimate variants without penalizing stylistic or structural divergence.
Future work will extend this framework to edit-based settings, linking fluency-oriented metrics with syntactic and grammatical fidelity.
Achieving this goal will require a unified, cross-linguistic grammatical annotation scheme, enabling consistent and interpretable comparisons across languages, corpora, and system architectures.

\section*{Limitations}

This study focuses on the theoretical formulation and analysis of fluency-based multi-reference evaluation, rather than on large-scale empirical benchmarking. 
We do not investigate correlations with human judgments or perform system-level comparisons across multiple GEC models, which are left for future work. 
In addition, while the proposed framework has been examined on several typologically diverse languages, broader cross-linguistic validation would be required to fully assess its generality and robustness across linguistic families and annotation conventions.

\section*{Data availability statement}
The multi-reference GLEU implementation will be released upon publication, in accordance with the licensing terms of the original GLEU repository at \url{https://github.com/cnap/gec-ranking}.

\section*{Acknowledgments}
Portions of this manuscript were reviewed with the assistance of AI-based tools, which were employed solely for grammar and consistency checking.


\appendix
\section{Mathematical formulation and properties of aggregation functions}
\label{appendix:math-formulation}

This appendix provides the formal definitions and analytical properties of the aggregation operators used in fluency-based multi-reference evaluation.

\paragraph{Formal definition}
Let $S$ denote the uncorrected source, $H$ the system hypothesis, and $\mathcal{R}=\{R_i\}_{i=1}^{n}$ the set of reference corrections.  
For each reference $R_i$, define the sentence-level fluency score as
\[
f_i = \text{GLEU}(S,H,R_i) \in [0,1]
\]
A general multi-reference fluency metric $\mathcal{F}$ can then be expressed as a mapping
\[
\mathcal{F}(f_1,\dots,f_n)
= \Phi(f_1,\dots,f_n)
\]
where $\Phi$ denotes an aggregation operator satisfying four desiderata:
\textit{boundedness}, \textit{monotonicity}, \textit{symmetry}, and \textit{consistency} with the single-reference case.

The four aggregation strategies instantiate $\mathcal{F}$ under distinct functional forms:
\[
\begin{aligned}
\mathcal{F}_{\textsc{select-best}} &= \max_i f_i\\
\mathcal{F}_{\textsc{simple-avg}} &= \tfrac{1}{n} \sum_i f_i\\
\mathcal{F}_{\textsc{weighted-avg}} &= \sum_i w_i f_i\\
\mathcal{F}_{\textsc{merged}} &= \text{GLEU}(S,H,R_{\cup})
\end{aligned}
\]
where $R_{\cup}$ denotes the union of all reference $n$-grams.
Each formulation inherits \textit{boundedness} and \textit{monotonicity} from GLEU while differing in \textit{symmetry} and sensitivity to reference variation.  
Formally, $\mathcal{F}_{\textsc{select-best}}$ provides a precision-oriented upper bound, $\mathcal{F}_{\textsc{merged}}$ a recall-oriented lower bound, and the averaging-based methods interpolate between these extremes.

\paragraph{Analytical properties}
We next formalize the analytical behavior of the aggregation operators.
Each property specifies a structural constraint or monotonicity condition that governs how multi-reference fluency scores evolve as the number and distribution of references vary.

\begin{property}[Oracle upper bound]
Let $f_i = \text{GLEU}(S,H,R_i)\in[0,1]$. For any $\{f_i\}_{i=1}^{n}$,
\[
\begin{aligned}
\mathcal{F}_{\textsc{select-best}}(f_1,\dots,f_n) &= \max_i f_i\\
\mathcal{F}_{\textsc{simple-avg}}(f_1,\dots,f_n) &\le \max_i f_i
\end{aligned}
\]
If $w_i\ge 0$ and $\sum_i w_i=1$, then
$\mathcal{F}_{\textsc{weighted-avg}}=\sum_i w_i f_i \le \max_i f_i$.
\end{property}

\begin{proofsketch}
The maximum is an upper bound by definition. Both simple and weighted averages are convex combinations, so they cannot exceed the maximum element.
\end{proofsketch}

\begin{interpretation}
\textsc{Select-best} represents an oracle upper bound, giving the highest possible score among references.  
Averaging methods are bounded above by the best reference and may underestimate system performance when reference diversity is high.
\end{interpretation}

\begin{property}[Monotonicity with respect to reference number]
Let $\mathcal{R}_n=\{R_i\}_{i=1}^{n}$ and $\mathcal{R}_{n+1}=\mathcal{R}_n\cup\{R_{n+1}\}$.  
Then
\[
\begin{aligned}
\mathcal{F}_{\textsc{select-best}}(\mathcal{R}_{n+1}) 
&\ge \mathcal{F}_{\textsc{select-best}}(\mathcal{R}_n) \\
\mathcal{F}_{\textsc{merged}}(\mathcal{R}_{n+1}) 
&\ge \mathcal{F}_{\textsc{merged}}(\mathcal{R}_n)
\end{aligned}
\]
\end{property}

\begin{proofsketch}
Adding a reference cannot reduce the maximum in \textsc{select-best}.  
For \textsc{merged}, the union $R_{\cup}^{(n+1)} = R_{\cup}^{(n)} \cup \text{ngrams}(R_{n+1})$ enlarges or preserves all rewarded $n$-grams, ensuring a non-decreasing score.
\end{proofsketch}

\begin{interpretation}
\textsc{Select-best} and \textsc{merged} satisfy the \textit{non-penalization principle}: adding more human references can only improve or maintain the score.
\end{interpretation}

\begin{property}[Limitation of averaging]
For \textsc{simple-average},
\[
\begin{aligned}
\mathcal{F}_{\textsc{simple-avg}}(\mathcal{R}_{n+1})
&= \frac{n}{n+1}\mathcal{F}_{\textsc{simple-avg}}(\mathcal{R}_n) \\
&\quad + \frac{1}{n+1} f_{n+1}.
\end{aligned}
\]
Hence, $\mathcal{F}_{\textsc{simple-avg}}(\mathcal{R}_{n+1}) \ge \mathcal{F}_{\textsc{simple-avg}}(\mathcal{R}_n)$
iff $f_{n+1}\ge \mathcal{F}_{\textsc{simple-avg}}(\mathcal{R}_n)$.
\end{property}

\begin{proofsketch}
Follows from the recursive definition of the arithmetic mean: the average increases only if the new score exceeds the current mean.
\end{proofsketch}

\begin{interpretation}
Simple averaging fails to guarantee monotonicity: adding a valid but divergent reference may lower the overall score, revealing a trade-off between fairness and robustness to variation.
\end{interpretation}

\begin{property}[Weighted averaging bounds and sensitivity]
With softmax weights $w_i = \mathrm{exp}(\tau f_i)/\sum_j \mathrm{exp}(\tau f_j)$ for $\tau>0$,
\[
\min_i f_i \le \sum_i w_i f_i \le \max_i f_i,
\]
and $\partial \mathcal{F}_{\textsc{weighted-avg}}/\partial f_k \ge 0$ for each $k$.  
Adding a low-scoring reference can slightly decrease the total score if it receives nonzero weight.
\end{property}

\begin{proofsketch}
A softmax-weighted sum is a convex combination.  
Gradients are nonnegative, but low-scoring references redistribute probability mass, which can lower the overall average.
\end{proofsketch}

\begin{interpretation}
Weighted averaging models graded human acceptability while maintaining boundedness.  
It captures how human evaluators tend to reward prototypical corrections more strongly than rare alternatives.
\end{interpretation}

\begin{property}[Non-penalization principle]
Among all strategies, \textsc{select-best} and \textsc{merged} satisfy non-penalization strictly, while average-based strategies do not.
\end{property}

\begin{interpretation}
This property summarizes the behavioral spectrum of multi-reference evaluation:
\textsc{select-best} corresponds to strict grading,
\textsc{merged} reflects inclusive evaluation,
and averaging-based strategies mediate graded fairness.
\end{interpretation}

\section{BLEU under multiple references}

To contextualize the \textsc{merged-$n$-grams} results, we briefly review how BLEU \citep{papineni-etal-2002-bleu} handles multiple references in machine translation.
Given a hypothesis $H$ and reference set $\{R_i\}_{i=1}^{n}$, BLEU defines its modified $n$-gram precision as the clipped overlap between hypothesis and reference $n$-grams:
\[
P_n =
\frac{
\sum_{g \in \text{ngrams}_n(H)}
\min\big(
\text{C}_H(g),
\max_i \text{C}_{R_i}(g)
\big)
}{
\sum_{g \in \text{ngrams}_n(H)} \text{C}_H(g)
}
\]
An $n$-gram in the hypothesis is rewarded if it appears in \emph{any} reference, but the matched count is capped by the maximum frequency across references rather than by their sum, preventing overcounting while still accounting for lexical diversity in human translations.

The brevity penalty (\textsc{BP}) is computed using the reference length $r^*$ closest to the candidate length $c$:
\[
\begin{aligned}
r^* &= \arg\min_{r_i} |r_i - c| \\
\text{BP} &=
\begin{cases}
1, & c > r^* \\
e^{(1 - r^*/c)}, & c \le r^*
\end{cases}
\end{aligned}
\]
The final BLEU score is
\[
\begin{aligned}
\text{BLEU}(H,\{R_i\})
&= \text{BP} \cdot
\mathrm{exp}\left(
\sum_{n=1}^{N} w_n \log P_n
\right) \\
&\text{where } w_n = \tfrac{1}{N},\; N = 4.
\end{aligned}
\]

\paragraph{Comparison with merged GLEU}
BLEU computes multi-reference precision by taking the \emph{maximum} count of each $n$-gram across references, forming a clipped union that prevents overcounting.
Our \textsc{merged-$n$-grams} GLEU instead constructs a true multiset union,
\[
R_{\cup} = \bigcup_i \text{ngrams}(R_i)
\quad
\text{C}_{R_{\cup}}(g) = \sum_i \text{C}_{R_i}(g)
\]
so that every attested human $n$-gram contributes in proportion to its occurrence across references.
This design shifts the metric from BLEU’s precision-oriented estimate of adequacy toward a recall-oriented representation of linguistic diversity.
Whereas BLEU stabilizes precision by limiting overlap to the most frequent realization, merged GLEU rewards the broader coverage of valid human corrections—an essential property for GEC, where lexical and syntactic variation reflects legitimate alternatives rather than noise.

\end{document}